\documentclass[11pt,a4paper]{article}
\usepackage{times,latexsym}
\usepackage{url}
\usepackage[T1]{fontenc}

\usepackage[acceptedWithA]{tacl2021v1}

\usepackage{xspace,mfirstuc,tabulary}

\newif\iftaclinstructions
\taclinstructionsfalse 
\iftaclinstructions
\renewcommand{\confidential}{}
\renewcommand{\anonsubtext}{(No author info supplied here, for consistency with
TACL-submission anonymization requirements)}
\newcommand{\instr}
\fi

\iftaclpubformat 

\else

\fi

\def\tv{t} 
\def\wv{w}    

\newcommand{\surp}[1]{\mathsf{S}( #1 )}
\newcommand{\condprob}[2]{\mathsf{P}( #1 \mid #2 )}

\usepackage{graphicx}
\usepackage{enumitem}

\title{Why Does Surprisal From Larger Transformer-Based Language Models Provide a Poorer Fit to Human Reading Times?}

\author{Byung-Doh Oh \\
  Department of Linguistics \\
  The Ohio State University \\
  \texttt{oh.531@osu.edu} \\\And
  William Schuler \\
  Department of Linguistics \\
  The Ohio State University \\
  \texttt{schuler.77@osu.edu} \\}

\date{}

\begin{document}
\maketitle
\begin{abstract}
This work presents a detailed linguistic analysis into why larger Transformer-based pre-trained language models with more parameters and lower perplexity nonetheless yield surprisal estimates that are less predictive of human reading times.
First, regression analyses show a strictly monotonic, positive log-linear relationship between perplexity and fit to reading times for the more recently released five GPT-Neo variants and eight OPT variants on two separate datasets, replicating earlier results limited to just GPT-2 \citep{ohetal22}.
Subsequently, analysis of residual errors reveals a systematic deviation of the larger variants, such as underpredicting reading times of named entities and making compensatory overpredictions for reading times of function words such as modals and conjunctions.
These results suggest that the propensity of larger Transformer-based models to `memorize' sequences during training makes their surprisal estimates diverge from humanlike expectations, which warrants caution in using pre-trained language models to study human language processing.
\end{abstract}

\section{Introduction}

Expectation-based theories of sentence processing \citep{hale01, levy08} postulate that processing difficulty is largely driven by how predictable upcoming linguistic material is given its context.
In cognitive modeling, predictability operationalized by information-theoretic surprisal \citep{shannon48} has been shown to be a strong predictor of behavioral and neural measures of processing difficulty \citep{dembergkeller08, smithlevy13, haleetal18, shainetal20}, providing empirical support for this position.
As language models (LMs) directly define a conditional probability distribution of a word given its context required for surprisal calculation, they have long been evaluated as surprisal-based cognitive models of sentence processing.

Recently, it was observed that surprisal from larger variants of the pre-trained GPT-2 LM \citep{radfordetal19} that have more parameters and achieve lower perplexity is less predictive of self-paced reading times and eye-gaze durations collected during naturalistic reading \citep{ohetal22}.
As the different variants of the pre-trained GPT-2 model share the primary architecture and training data, this offers an especially strong counterexample to previous work that showed a negative relationship between LM perplexity and predictive power of surprisal estimates \citep{goodkindbicknell18, haoetal20, wilcoxetal20}.
More broadly, this observation also contradicts the recent `larger is better' trend of the NLP community, leaving open the question of why larger LMs perform worse.
However, the \citet{ohetal22} results were part of a follow-up analysis in support of a separate claim about parser surprisal that only examined four model variants, so the results were not tested for statistical significance or extensively explored.

The current work fills that gap by conducting a detailed linguistic analysis of the positive relationship between LM perplexity and predictive
power of surprisal estimates.
First, the robustness of the trend observed in \citet{ohetal22} was examined by reproducing their results and additionally evaluating surprisal estimates from different families of Transformer-based LMs \citep[GPT-Neo, OPT;][]{blacketal21, blacketal22, wangkomatsuzaki21, zhangetal22} on their ability to predict human reading times.
Results from regression analyses show a strictly monotonic, positive log-linear relationship between LM perplexity and fit to reading times for the five GPT-Neo variants and eight OPT variants on two separate datasets, which provides firm empirical support for this trend.
Subsequently, to provide an explanation for this positive relationship, residual errors from the regression models were analyzed with a focus on identifying linguistic phenomena that surprisal from larger variants accounts for less accurately compared to their smaller counterparts.
The results show that regression models with surprisal predictors from GPT-2, GPT-Neo, and OPT models generally underpredict reading times at nouns and adjectives, and that the degree of underprediction increases along with model size.
This indicates that the poorer fit to human reading times achieved by surprisal estimates from larger Transformer-based LMs is primarily driven by their characteristic of assigning lower surprisal values to open-class words, which may be accurately predicted by extensive domain knowledge gleaned from large sets of training examples that are not available to humans.
This suggests that as Transformer-based LMs get larger, they may be problematic for cognitive modeling because they are trained with non-human learning objectives and different inductive biases on vast quantities of Internet text.\footnote{All code used in this work is available at: \url{https://github.com/byungdoh/llm_surprisal}.}

\section{Related Work} \label{sec:bkgd}
In previous studies, surprisal estimates from several well-established types of LMs, including $n$-gram models, Simple Recurrent Networks \citep{elman91}, Gated Recurrent Unit networks \citep[GRU;][]{choetal14}, and Long Short-Term Memory networks \citep[LSTM;][]{Hochreiter1997}, have been compared against behavioral measures of processing difficulty \citep[e.g.][]{smithlevy13, goodkindbicknell18, aurnhammerfrank19rnn}.
Recently, as Transformer-based \citep{vaswanietal17transformer} models have dominated many NLP tasks, both large pre-trained and smaller `trained-from-scratch' Transformer-based LMs have also been evaluated as models of processing difficulty \citep{wilcoxetal20, haoetal20, merkxfrank21, schrimpfetal21}.

A consistent finding that emerged out of these studies is that better LMs are also more predictive models of comprehension difficulty, or in other words, there is a negative correlation between LM perplexity and fit to human reading times.
\citet{goodkindbicknell18} compared surprisal estimates from a set of $n$-gram and LSTM LMs and observed a negative linear relationship between perplexity and regression model fit.
\citet{wilcoxetal20} evaluated $n$-gram, LSTM, Transformer, and RNNG \citep{dyeretal16} models and replicated the negative relationship, although they note a more exponential relationship at certain intervals.
\citet{merkxfrank21} provided further support for this trend using GRU and Transformer models with different numbers of layers.\footnote{Although counterexamples to this trend have been noted, they were based on comparisons of LMs and incremental parsers that were trained on different data \citep{ohetal21acl} or evaluation on Japanese, which has a different syntactic head-directionality than English \citep{kuribayashietal21}.}

However, \citet{ohetal22} observed a directly contradictory relationship to this using surprisal estimates from pre-trained GPT-2 models \citep{radfordetal19}.
Using self-paced reading times from the Natural Stories Corpus \citep{futrelletal21} and go-past durations from the Dundee corpus \citep{kennedyetal03}, the authors calculated the increase in log-likelihood ($\Delta$LL) to a baseline linear-mixed effects (LME) model due to including a surprisal predictor.
Their results showed that surprisal from the largest \textit{XL} variant made the smallest contribution to regression model fit, followed by the smaller \textit{Large}, \textit{Medium}, and \textit{Small} variants in that order, revealing a robust positive correlation between LM perplexity and predictive power of surprisal estimates.
The same trend was replicated when unigram surprisal was included in the baseline, as well as when spillover effects were controlled for through the use of continuous-time deconvolutional regression \citep[CDR;][]{shainschuler21}.

Moreover, recent work has shown that surprisal from neural LMs generally tends to underpredict human reading times of both targeted constructions and naturalistic text.
For instance, \citet{vanschijndellinzen21} and \citet{arehallietal22} observed that surprisal from neural LMs severely underpredicts the magnitude of garden-path effects demonstrated by human subjects.
Additionally, \citet{hahnetal22} showed that surprisal from the pre-trained GPT-2 model fails to accurately predict the increase in reading times at the main verb of deeply embedded sentences.
\citet{kuribayashietal22} also demonstrated that neural LMs yield surprisal estimates that underpredict naturalistic reading times of English and Japanese text compared to those from neural LMs that have a recency bias implemented as limited access to the previous context.

\section{Main Experiment: Predictive Power of Language Model Surprisal Estimates} \label{sec:expl}

In order to examine whether the positive correlation observed by \citet{ohetal22} and others generalizes to larger Transformer-based models, surprisal predictors from different variants of the GPT-2, GPT-Neo, and OPT LMs were evaluated on self-paced reading times from the Natural Stories Corpus \citep{futrelletal21} and go-past eye-gaze durations from the Dundee Corpus \citep{kennedyetal03}.

\subsection{Response Data}
The Natural Stories Corpus contains data from 181 subjects that read 10 naturalistic English stories that consist of a total of 10,245 tokens.
The reading times were filtered to remove observations for sentence-initial and sentence-final words, observations from subjects who answered three or fewer comprehension questions correctly, and observations shorter than 100 ms or longer than 3000 ms, which resulted in a total of 770,102 observations.
The Dundee Corpus contains data from 10 subjects that read 67 newspaper editorials that consist a total of 51,501 tokens.
The durations were filtered to remove observations for unfixated words, words following saccades longer than four words, and words at sentence-, screen-, document-, and line-starts and ends.
This resulted in a total of 195,507 observations.

Both datasets were subsequently partitioned into an exploratory set and a held-out set of roughly equivalent sizes.\footnote{This partitioning was conducted based on the sum of subject ID and sentence ID, resulting in each subject-by-sentence combination remaining intact in one partition.}
This partitioning allows regression model selection (e.g.~making decisions about random effects structure) and exploratory analyses to be conducted on the exploratory set and a single statistical significance test to be conducted on the held-out set, thereby obviating the need for multiple trials correction.
This resulted in an exploratory set of 384,905 observations and a held-out set of 385,197 observations for the Natural Stories Corpus and an exploratory set of 98,115 observations and a held-out set of 97,392 observations for the Dundee Corpus.~All observations were log-transformed prior to model fitting.

\subsection{Predictors}

Surprisal estimates calculated from four different variants of GPT-2 models \citep{radfordetal19} were used in \citet{ohetal22}.
In addition to GPT-2 surprisal, this experiment also evaluates surprisal estimates from five variants of GPT-Neo models \citep{blacketal21, blacketal22, wangkomatsuzaki21}\footnote{Technically, the two largest variants are GPT-J and GPT-NeoX models respectively, both of which have minor architectural differences from the GPT-Neo models. However, given that they share the same training data, they were considered to belong to the same family as the GPT-Neo models.} and eight variants of OPT models \citep{zhangetal22}.\footnote{The largest variant of the OPT model, which has about 175 billion parameters, was not used in this work due to constraints in computational resources.}
All of these LMs are decoder-only autoregressive Transformer-based models whose variants mainly differ in their capacity.
The model capacities of the three LM families are summarized in Table \ref{tab:params}.

\begin{table}[t!]
    \centering
    \footnotesize
    \begin{tabular}{lrrrr}
    Model & \#L & \#H & $d_{model}$ & Parameters \\ \hline
    GPT-2 Small & 12 & 12 & 768 & $\sim$124M \\
    GPT-2 Medium & 24 & 16 & 1024 & $\sim$355M \\
    GPT-2 Large & 36 & 20 & 1280 & $\sim$774M \\
    GPT-2 XL & 48 & 25 & 1600 & $\sim$1558M \\ \hline
    GPT-Neo 125M & 12 & 12 & 768 & $\sim$125M \\
    GPT-Neo 1300M & 24 & 16 & 2048 & $\sim$1300M \\
    GPT-Neo 2700M & 32 & 20 & 2560 & $\sim$2700M \\
    GPT-J 6B & 28 & 16 & 4096 & $\sim$6000M \\
    GPT-NeoX 20B & 44 & 64 & 6144 & $\sim$20000M \\ \hline
    OPT 125M & 12 & 12 & 768 & $\sim$125M \\
    OPT 350M & 24 & 16 & 1024 & $\sim$350M \\
    OPT 1.3B & 24 & 32 & 2048 & $\sim$1300M \\
    OPT 2.7B & 32 & 32 & 2560 & $\sim$2700M \\
    OPT 6.7B & 32 & 32 & 4096 & $\sim$6700M \\
    OPT 13B & 40 & 40 & 5120 & $\sim$13000M \\
    OPT 30B & 48 & 56 & 7168 & $\sim$30000M \\
    OPT 66B & 64 & 72 & 9216 & $\sim$66000M \\
    \end{tabular}
    \caption{Model capacities of LM families whose surprisal estimates were examined in this work. \#L, \#H, and $d_{model}$ refers to number of layers, number of attention heads, and embedding size respectively.}
    \label{tab:params}
\end{table}

Each story of the Natural Stories Corpus and each article of the Dundee Corpus was tokenized according to the three models' respective byte-pair encoding \citep[BPE;][]{sennrichetal15} tokenizer and was provided to each model variant to calculate surprisal estimates.
In cases where each story or article did not fit into a single context window for the LMs, the second half of the previous context window served as the first half of a new context window to calculate surprisal estimates for the remaining tokens.
In practice, most stories and articles fit completely within two context windows for the GPT-2 models that have a context size of 1,024 tokens, and within one context window for the GPT-Neo and OPT models that have a context size of 2,048 tokens.
Additionally, when a single word $\wv_{\tv}$ was tokenized into multiple subword tokens, negative log probabilities of subword tokens corresponding to $\wv_\tv$ were added together to calculate $\surp{ \wv_{\tv} } = -\log\condprob{ \wv_\tv }{ \wv_{1..\tv-1} }$.

\subsection{Regression Modeling} \label{sec:reg}
Subsequently, following the methods of \citet{ohetal22}, a `baseline' LME model that contains baseline predictors capturing low-level cognitive processing and seventeen `full' LME models that contain the baseline predictors and each LM surprisal predictor were fit to the exploratory set of self-paced reading times and go-past durations using \texttt{lme4} \citep{batesetal15}.
The baseline predictors include word length measured in characters and index of word position within each sentence (both self-paced reading and eye-tracking), as well as saccade length and whether or not the previous word was fixated (eye-tracking only).

All predictors were centered and scaled prior to model fitting, and the LME models included by-subject random slopes for all fixed effects as well as random intercepts for each subject and each word type.
Additionally, for self-paced reading times collected from 181 subjects, a random intercept for each subject-sentence interaction was included.
For eye-gaze durations collected from a much smaller number of 10 subjects, a random intercept for each sentence was included.

After the regression models were fit, the $\Delta$LL values were first calculated for each regression model by subtracting the log-likelihood of the baseline model from that of a full regression model.
Moreover, to examine the trend between LM perplexity and predictive power of surprisal estimates, the perplexity of each LM variant was calcuated on the two corpora.

\begin{figure}[t!]
    \centering
    \includegraphics[width=0.48\textwidth]{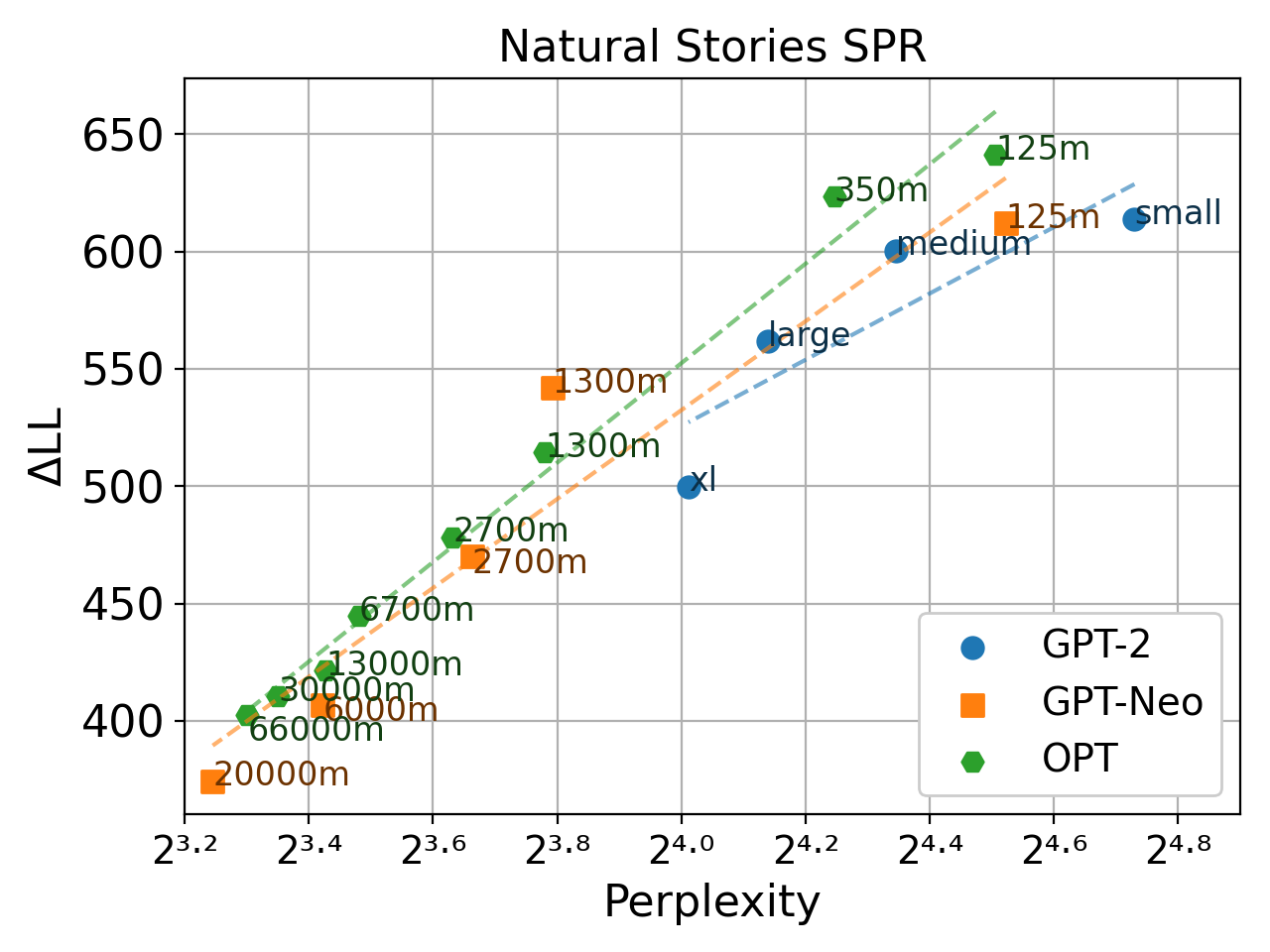}
    \includegraphics[width=0.48\textwidth]{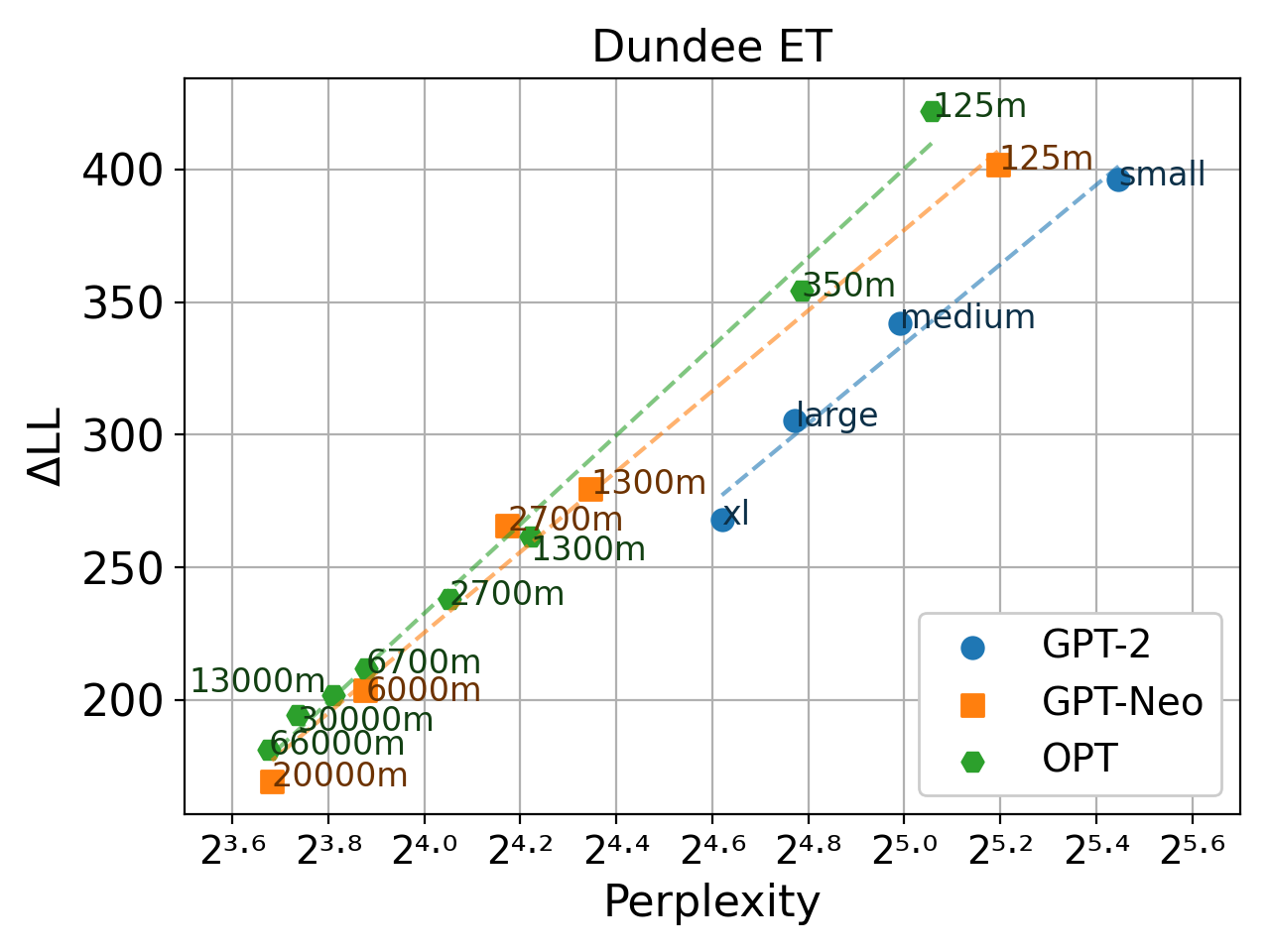}
    \caption{Perplexity measures from each LM variant, and improvements in regression model log-likelihood from including each surprisal estimate on the exploratory set of Natural Stories (top) and Dundee data (bottom). Dotted lines indicate the least-squares regression line for each LM family.}
    \label{fig:replication}
\end{figure}

\subsection{Results} \label{sec:results}

The results in Figure \ref{fig:replication} show that surprisal from the smallest variant (i.e.~GPT-2 Small, GPT-Neo 125M, and OPT 125M) made the biggest contribution to regression model fit on both self-paced reading times and eye-gaze durations for the three LM families.
More notably, surprisal estimates from larger LM variants within each family yielded strictly poorer fits to reading times, robustly replicating the trend observed by \citet{ohetal22}.
Interestingly, the three LM families also seem to demonstrate a strong log-linear relationship between perplexity and $\Delta$LL, as can be seen by the least-squares regression lines.
All regression lines had a slope significantly greater than 0 at $p<0.05$ level according to a one-tailed $t$-test, with the exception of the regression line for GPT-2 on Natural Stories ($p=0.07$). 
This trend is highly significant overall by a binomial test (five results with $p<0.05$ out of six trials), and directly contradicts the findings of recent studies that report a negative correlation between LM perplexity and predictive power of surprisal estimates.

Additionally, comparison of the GPT-2 models and OPT models of similar model capacities (i.e.~Small-125M, Medium-350M) shows that the OPT models generally both achieve lower perplexity and yield surprisal estimates that are more predictive of human reading times.
Given the high similarity in model architecture between the two LMs, this trend seems to be due to the difference in the training data that were used.
The most notable difference between the two training datasets is in their size, with the training set for GPT-2 estimated to be about 15B tokens and that for OPT estimated to be about 180B tokens \citep{thompson22}.
However, the GPT-Neo models trained on about 247B tokens show no improvement over the OPT models, yielding a mixed picture.
These results suggest that beyond a certain level, the quantity of training data may play a secondary role to the number of model parameters in capturing humanlike expectations.

\section{Post-hoc Analysis: Linguistic Phenomena Underlying the Trend}

In order to provide an explanation for the trend observed in Section \ref{sec:expl}, the residual errors from the regression models were analyzed to identify data points that surprisal from larger LM variants accounted for less accurately compared to their smaller counterparts.
For this analysis, a special emphasis was placed on identifying subsets of data points where surprisal from larger LM variants deviated more drastically from humanlike processing difficulty.

\subsection{Calculation of Residual Errors}
The seventeen LME models that contain each of the LM surprisal predictors described in Section \ref{sec:reg} were used to generate predictions for all data points  in the exploratory set of both self-paced reading times and go-past durations.\footnote{The post-hoc analysis focused on the exploratory set, as the held-out set is reserved for statistical significance testing.}
Subsequently, the predictions were subtracted from the target values to calculate the residual errors for each of the seventeen regression models.

\begin{figure}[ht!]
    \centering
    \includegraphics[width=0.48\textwidth]{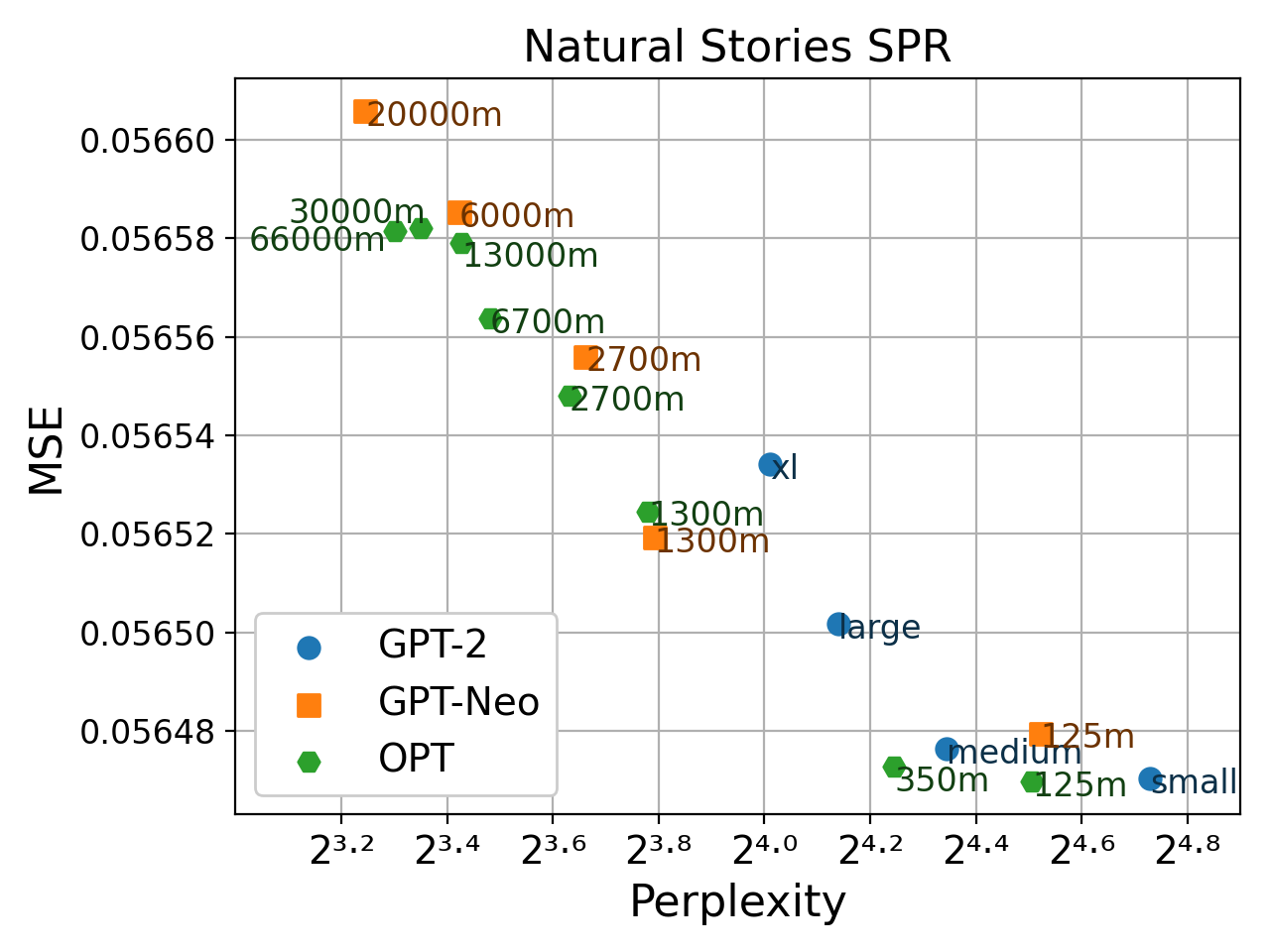}
    \includegraphics[width=0.48\textwidth]{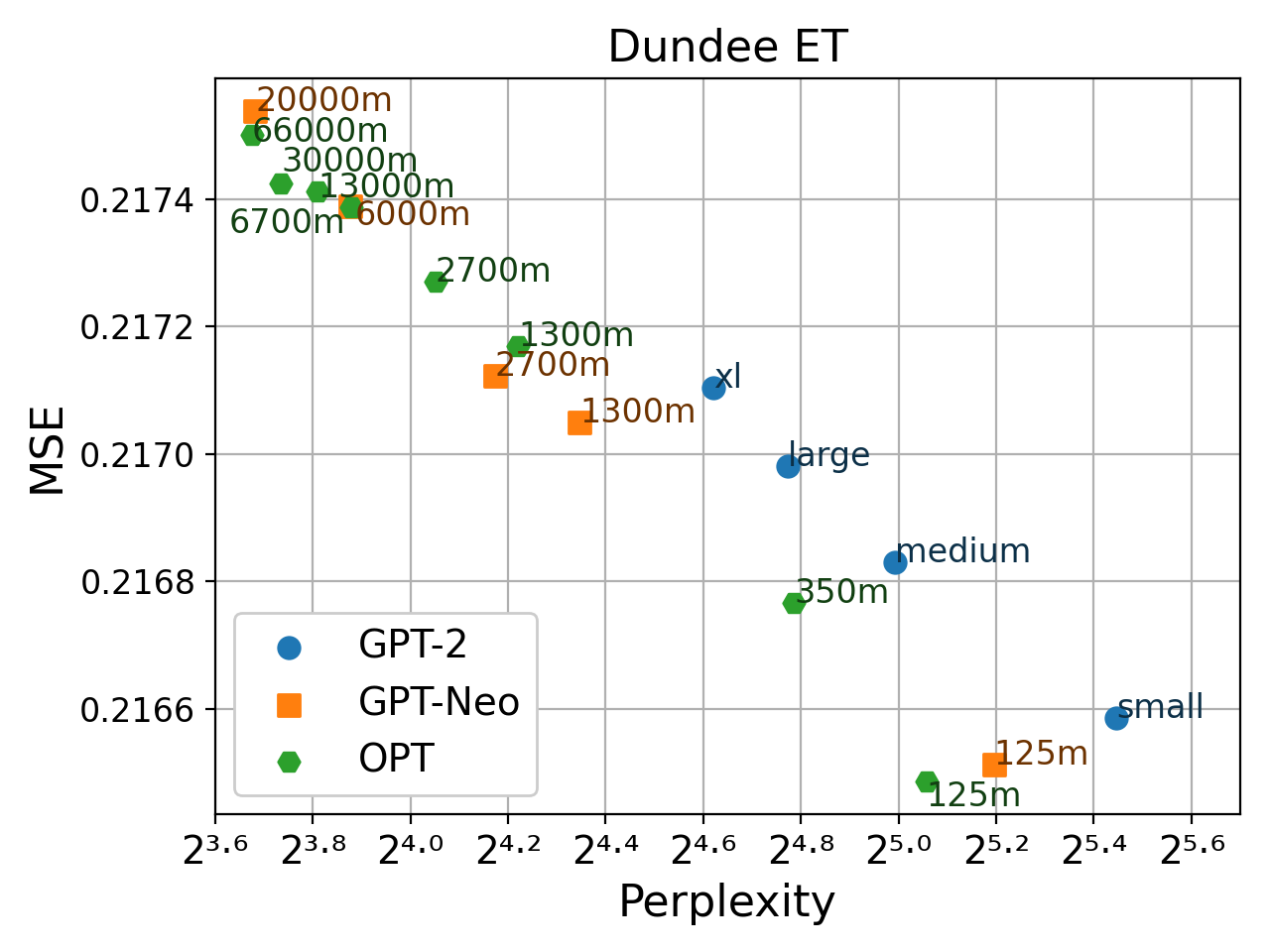}
    \caption{Perplexity measures from each LM variant, and mean squared errors of regression models that include each surprisal estimate on the exploratory set of Natural Stories (top) and Dundee data (bottom). Note that the larger values of MSE correspond to smaller values of log-likelihood in Figure \ref{fig:replication}.}
    \label{fig:mse}
\end{figure}

However, a preliminary analysis of the LME models fitted to the Dundee Corpus revealed a discrepancy between model likelihood and mean squared error (MSE), where the regression models with higher likelihoods achieved similar MSEs to those with lower likelihoods.
This is because the \texttt{lme4} package \citep{batesetal15} minimizes the \textit{penalized} residual sum-of-squares, which includes a Euclidean norm penalty on the spherical component of the random effects variables.
In other words, an LME model can achieve higher likelihood than another if it can achieve similar MSE using less expressive random effects variables that have lower variance.

An inspection of the fitted random effects variables revealed that the by-word intercept was mostly responsible for the discrepancy between likelihood and MSE for the LME models fitted to the Dundee Corpus.
More specifically, the LME models with surprisal estimates from larger LM variants had systematically higher variance for the by-word intercept, which allowed them to achieve similar MSEs at the cost of an increased penalty for the random effects variables.
In order to control for this confound and bring model likelihood and MSE closer together, the seventeen LME models were fitted again to both corpora with the by-word random intercepts removed.
Since the goal of this analysis was to identify data points that are responsible for the positive correlation between LM perplexity and fit to human reading times, it was thought that removing the by-word random intercepts would also yield a clearer picture with regard to words on which the surprisal estimates from larger LMs fall especially short.

The MSEs plotted in Figure \ref{fig:mse}, which generally replicate the inverse trend of $\Delta$LLs in Figure \ref{fig:replication}, show that the removal of by-word random intercepts brought model likelihoods and MSEs closer.
The residual errors from these newly fitted regression models were subsequently analyzed.

\subsection{Annotation of Data Points} \label{sec:annot}

In order to guide the identification of linguistic phenomena underlying the trend observed in Section \ref{sec:results}, each data point in both corpora was associated with various word- and sentence-level properties that are thought to influence real-time processing.
These properties were derived from the manually annotated syntactic tree structures of both corpora from \citet{shainetal18:lincr}.

Word-level properties reflect characteristics of the word that generally hold regardless of the surrounding context:
\begin{itemize}[leftmargin=*]
    \setlength\itemsep{0em}
    \item Part-of-speech: the syntactic category of each word from a generalized categorial grammar annotation scheme \citep{nguyenetal12,shainetal18:lincr}.~As these categories are defined in terms of primitive types (e.g.~verbs and nouns) and type-combining operators (e.g.~unsatisfied preceding and succeeding arguments), they make more fine-grained distinctions in terms of linguistic subcategorization.
    \item Named entities: a binary variable for whether or not the word is part of a proper name. Since words at the beginning of sentences were excluded from regression modeling, capitalization reliably identified such named entities.\footnote{Words like the pronoun \textit{I} and names of fictional characters that appeared in the Natural Stories Corpus were manually excluded afterwards.}
\end{itemize}

Sentence-level properties capture the syntactic structure of sentences, either in terms of dependencies or hierarchical phrases:
\begin{itemize}[leftmargin=*]
    \setlength\itemsep{0em}
    \item Dependency Locality Theory \citep[DLT;][]{gibson00} cost: DLT posits that the construction of backward-looking dependencies between words (e.g.~between a verb and its subject) incurs an `integration' cost driven by memory retrieval operations.
    This cost is thought to be proportional to the length of the dependency in terms of the number of intervening discourse referents, which are operationalized as any noun or finite verb in this work.
    \item Left-corner parsing \citep{johnsonlaird83}: A left-corner parser incrementally derives phrasal structures from a series of lexical match and grammatical match decisions at every word.\footnote{See e.g.~\citet{ohetal22} for a more detailed definition of left-corner parsing models.}
    These two decisions allow center-embedded constituents to be distinguished from non-embedded constituents.
    Additionally, the grammatical decision results in expectations about the upcoming syntactic category, which allows words before complete constituents (e.g.~words before sentential clauses) to be identified.
\end{itemize}

This annotation allowed the data points in each corpus to be subsetted, which subsequently helped identify where surprisal from the larger LM variants deviated further from humanlike processing.

\subsection{Iterative Slope-Based Analysis of Residual Errors}
Subsequently, based on the properties annotated in Section \ref{sec:annot}, subsets of data points that strongly drive the trend in Figure \ref{fig:mse} were identified.
To this end, the linear relationship between log perplexity and MSEs was used; subsets of data points that drive the general trend should show larger differences in MSE between regression models, or in other words, have negative slopes that are steeper than the corpus-level slope.

Based on this idea, for every corpus-LM combination (i.e.~\{Natural Stories, Dundee\} $\times$ \{GPT-2, GPT-Neo, OPT\}), a least-squares regression line was fitted between corpus-level log perplexity and MSEs of each subset defined by the properties outlined in Section \ref{sec:annot}.
Subsequently, the subset with the steepest negative slope was identified.
After excluding the identified subset, the above procedure was repeated to identify a new subset that showed the next strongest effect.
For this analysis, only subsets that contained more than 1\% of the data points in each corpus were considered at each iteration.\footnote{This criterion amounts to $>$3,849 data points for Natural Stories and $>$981 data points for Dundee at the first iteration. Although this may seem like a lenient criterion, this was necessary to examine phenomena that lie at the long tail of the Zipfian distribution of word frequencies.}

\begin{figure*}[ht!]
    \centering
    \includegraphics[width=0.98\textwidth]{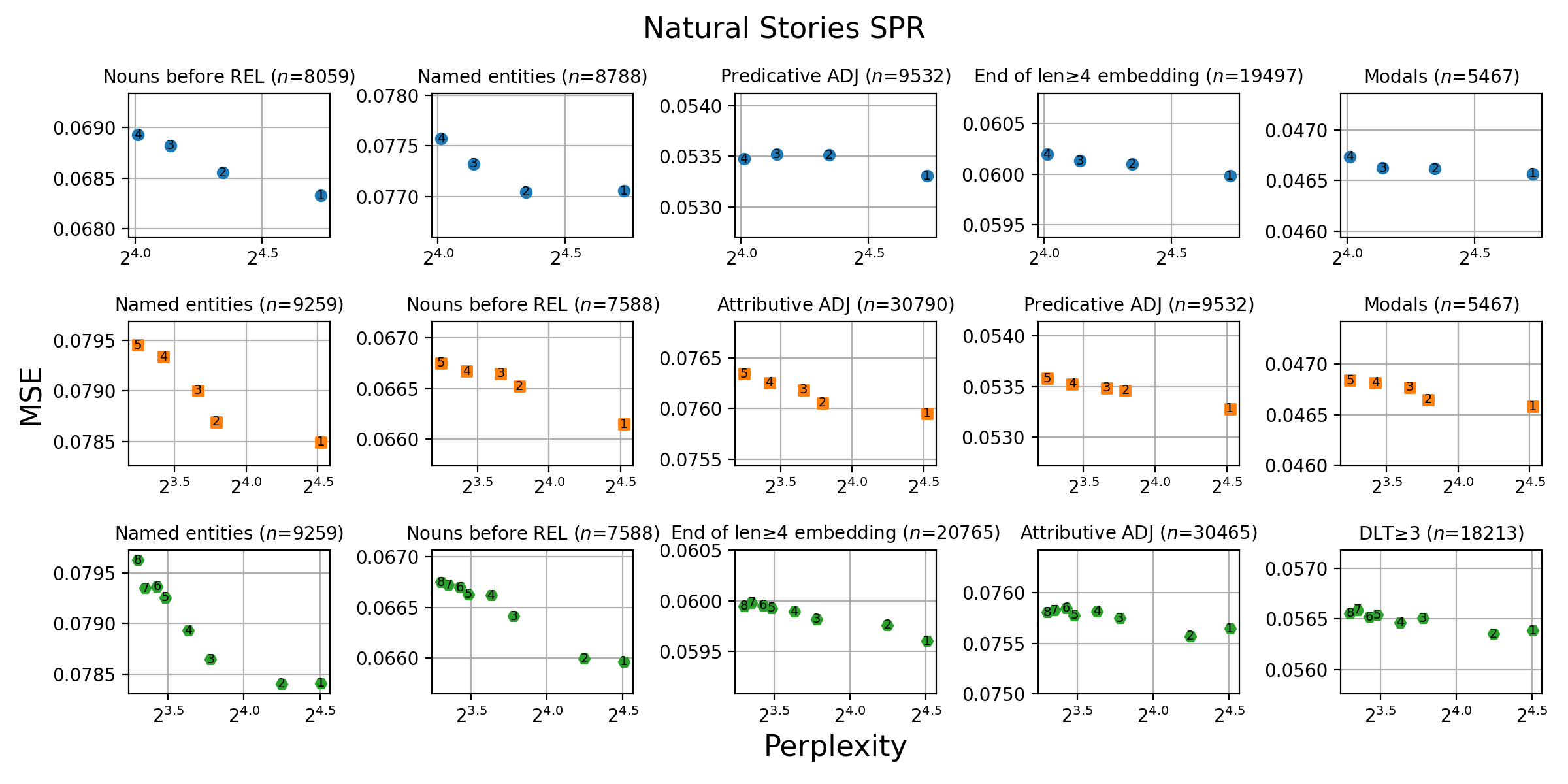}
    \includegraphics[width=0.98\textwidth]{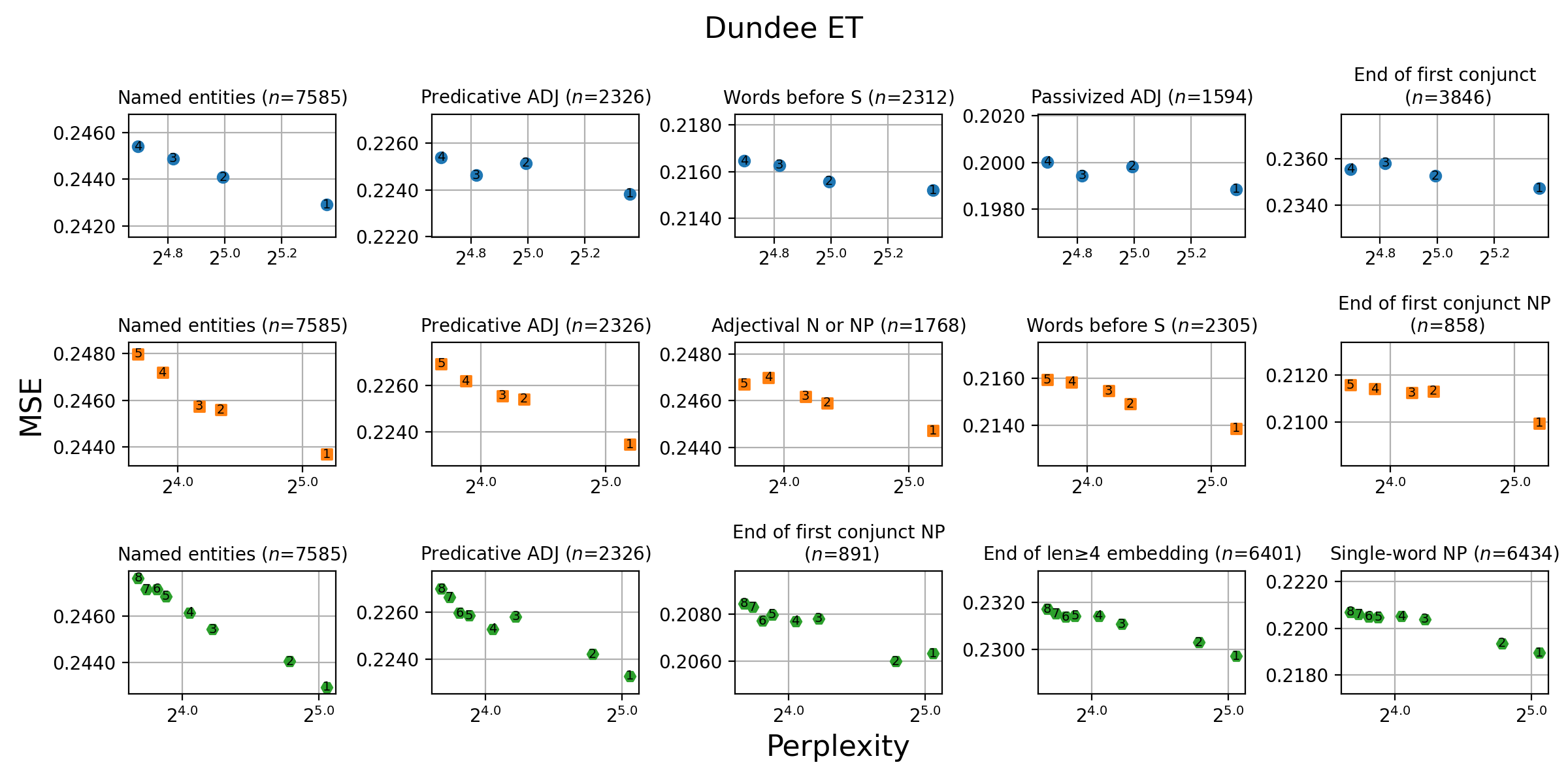}
    \caption{Corpus-level perplexity measures from each GPT-2, GPT-Neo, and OPT model variant (top, middle, and bottom rows respectively), and mean squared errors of regression models that include each surprisal estimate on the top five subsets (columns ordered from left to right) of Natural Stories self-paced reading data (top panel) and Dundee eye-tracking data (bottom panel). The ordered labels represent LM variants of different sizes, with `1' representing the smallest variant. ADJ: adjective, N: noun, NP: noun phrase, REL: relativizer, S: sentential clause.}
    \label{fig:ppl_mse}
\end{figure*}

\begin{figure*}[ht!]
    \centering
    \includegraphics[width=0.98\textwidth]{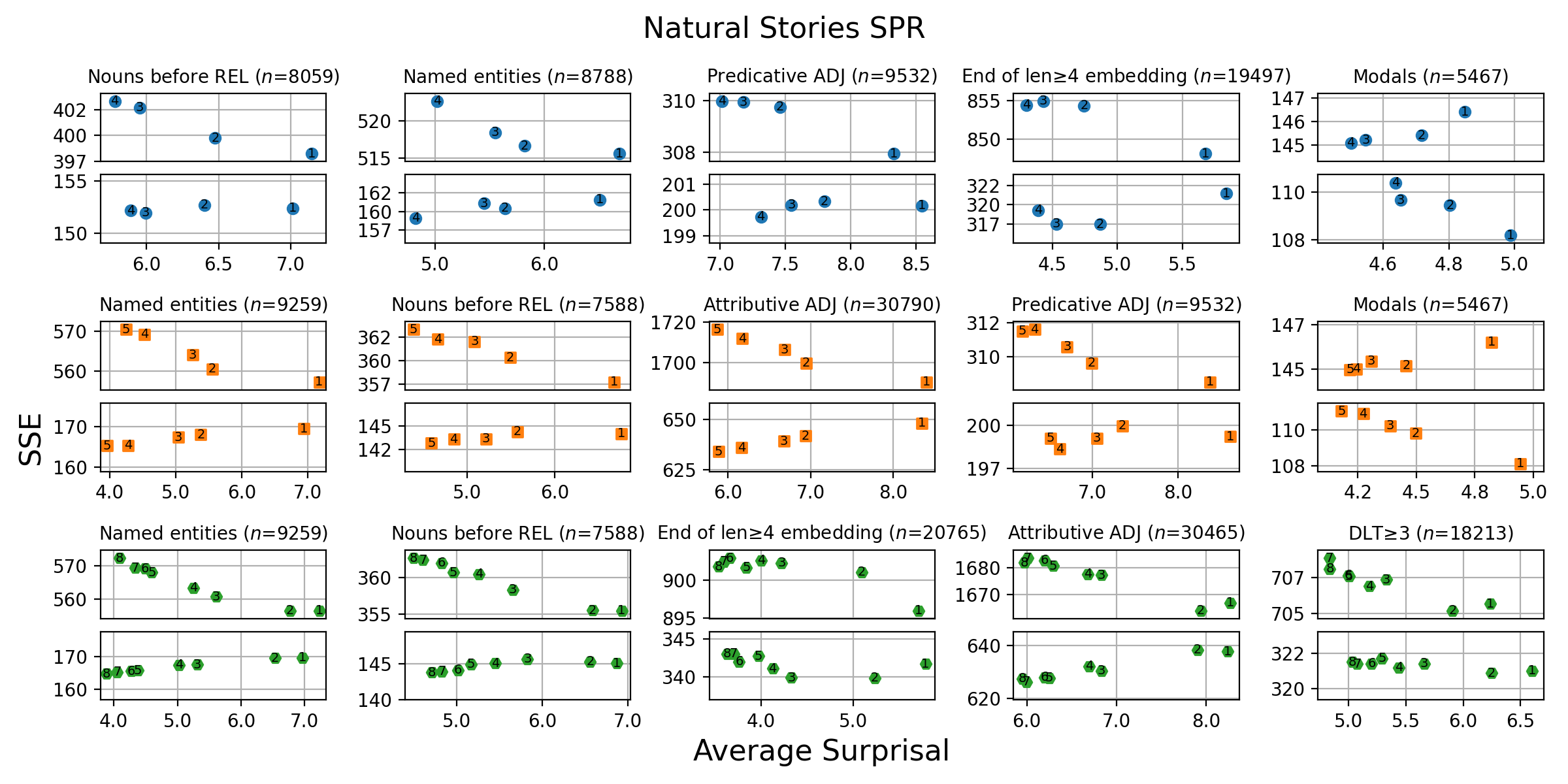}
    \includegraphics[width=0.98\textwidth]{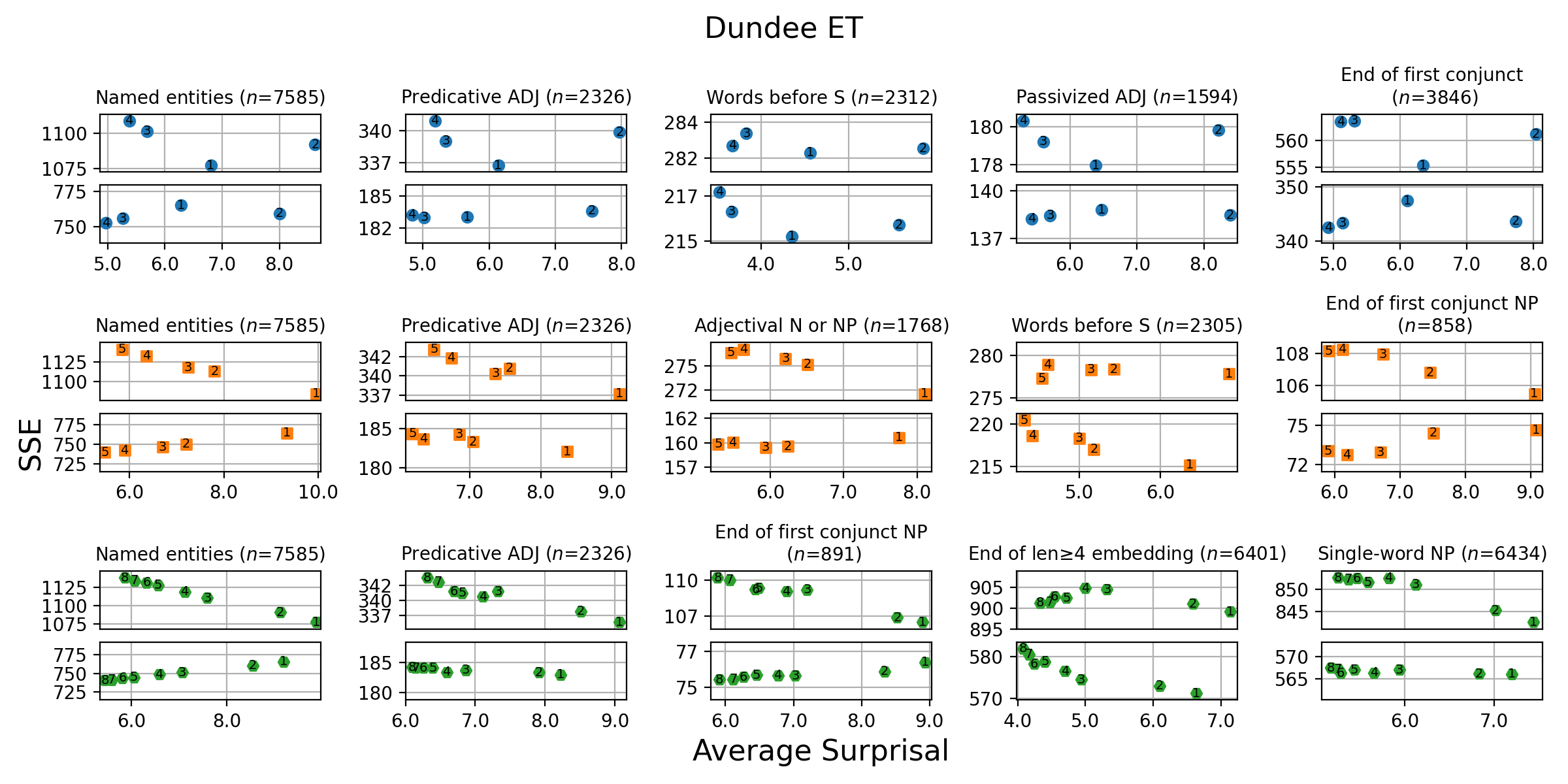}
    \caption{Average surprisal from each GPT-2, GPT-Neo, and OPT model variant, and sum of squared errors of regression models that include each surprisal estimate on the top five subsets of Natural Stories self-paced reading data and Dundee eye-tracking data. The top and bottom subplots of each row represent values from underpredicted and overpredicted data points respectively.}
    \label{fig:overunder}
\end{figure*}

Additionally, once the subsets of interest were identified, the data points in each subset were further separated according to whether the regression model underpredicted or overpredicted the target reading times.
In order to identify whether the trend of MSEs is driven primarily by systematic underprediction or overprediction of reading times, the average surprisal from each LM variant and the sum of squared errors (SSE) were calculated for each subset.
SSEs instead of MSEs were analyzed because different regression models had different numbers of underpredicted vs.~overpredicted data points, and because points close to 0 can distort the MSEs and obscure the overall trend of mispredictions. 

\subsection{Results}

The results in Figure \ref{fig:ppl_mse} show that on each corpus, similar subsets were identified as driving the trend of MSEs across different LM families.
On the Natural Stories Corpus, these subsets were primarily determined by the word's syntactic category, such as named entity nouns, nouns before relativizers, attributive and predicative adjectives, and modals.
The top subsets of the Dundee Corpus were similarly determined by syntactic category, such as named entity nouns, predicative and passivized adjectives, and single-word noun phrases (e.g.~pronouns).
Subsets defined by the syntactic structure of sentences were less commonly identified from the Natural Stories Corpus, with ends of center-embedded constituents spanning four or more words and words with high DLT costs emerging.
From the Dundee Corpus, ends of center-embedded constituents, ends of first conjunct constituents (both overall and noun phrases specifically), and beginnings of adjectival noun phrases (e.g.~\textit{family} in \textit{a family size pack}) were identified.
Additionally, words preceding a sentential clause were identified, which corresponded to conjunctions and ends of adjuncts.
On most of these subsets, the MSEs of each regression model were higher than those on the entire corpus, which indicates that the misprediction of reading times that pre-trained LM surprisal already has difficulty modeling is exacerbated as the models get larger.
Subsets such as modals of Natural Stories and first conjunct NP endings of Dundee are counterexamples to this general trend.

The average surprisal values\footnote{Since perplexity is equivalent to exponentiated average surprisal, the average surprisal values for each subset are roughly comparable to LM perplexity of e.g.~Figure \ref{fig:mse}. However, caution is warranted as these values are calculated over data points of reading times instead of tokens.} and SSEs from underpredicted and overpredicted data points in Figure \ref{fig:overunder} shed more light on the direction and magnitude of mispredictions from each regression model.
For example, on the subset of named entities, which emerged as the top two subsets across all corpus-by-LM combinations, the larger LM variants show systematically higher SSEs due to underprediction.
This strong discrepancy highlights a mismatch between human sentence processing and language modeling; named entity terms (e.g.~\textit{Elvis Presley}) have been shown to incur increased processing times compared to their common noun counterparts (e.g.~\textit{a singer}) due to various semantic associations that are retrieved \citep{proverbioetal01, wangetal13}.
In contrast, for language models, named entity terms that typically consist of multiple tokens have high mutual information, making it easy for them to accurately predict subsequent tokens given the first (e.g.~\textit{Presley} given \textit{Elvis}), resulting in especially lower surprisal estimates for larger LM variants.

Similarly, across the two corpora and three LM families, the trend of MSEs for other nouns as well as adjectives appears to be consistently driven by more severe underpredictions from regression models containing surprisal estimates from larger LM variants.
On these subsets, the difference in average surprisal values between the smallest and largest LM variants was typically above 2 bits, which is larger than the difference in log perplexity (i.e.~corpus-level average surprisal, Figure \ref{fig:mse}) between these variants.
This indicates that these subsets represent words that the larger LM variants predict especially accurately, which results in low surprisal estimates that deviate from human reading times.

In contrast, the subset of modals on the Natural Stories Corpus identified for the GPT-2 and GPT-Neo models shows a completely opposite trend in which more severe overpredictions drive the overall trend of MSEs.
This seems to be more due to the difference in the estimated regression coefficients rather than the difference in the LM surprisal estimates themselves.
The average surprisal values on this subset show that their difference between the smallest and largest variants is less than 1 bit, which indicates that the LM variants are making more similar predictions about modals.
However, since surprisal predictors from larger LM variants are generally smaller in magnitude, the regression models assign them higher coefficients in order to predict reading times, resulting in a systematic overprediction given surprisal predictors of similar values.
This also explains the trend observed for words preceding a sentential clause on the Dundee Corpus, which mainly consisted of conjunctions.

Finally, while they were less common, subsets based on syntactic complexity were also identified as driving the differential fit to reading times.
On the Natural Stories Corpus, a systematic underprediction of regression models with OPT surprisal was observed on words with a DLT cost of greater than or equal to three.
These words mainly consist of nouns and finite verbs that complete long-distance dependencies.
While finite verbs in general were not identified as subsets that showed a strong effect, it is likely that the increased reading times caused by the construction of long-distance dependencies made the underpredictions more salient.
Ends of center-embedded constituents of length greater than or equal to four were also identified, which typically corresponded to nouns and adjectives.
On the Natural Stories Corpus, the trend of more severe underpredictions driving the effect is consistent with other noun and adjective subsets.
However, on the Dundee Corpus, overpredictions seem to be responsible for the overall trend in this subset, which may hint at subtle differences in how syntactic complexity is manifested in self-paced reading times and eye-gaze durations.

Taken together, these results indicate that the poorer fit to human reading times achieved by surprisal estimates from larger Transformer-based language models is primarily driven by their characteristic of assigning lower surprisal values to open-class words like nouns and adjectives, which may be accurately predicted by extensive domain knowledge gleaned from large sets of training examples that are not available to humans.
In other words, the extra parameters of the larger LM variants may be improving predictions of such words in a way that is beyond human ability.

\section{Discussion and Conclusion}

This work presents results using multiple large pre-trained LMs showing that larger variants with more parameters and better next-word prediction performance (i.e.~lower perplexity) nonetheless yield surprisal estimates that are less predictive of human reading times (i.e.~smaller contribution to regression model fit), corroborating and expanding upon earlier results based on the GPT-2 LM \citep{ohetal22}.

First, in order to examine the generalizability of this trend, surprisal estimates from five variants of the GPT-Neo LM and eight variants of the OPT LM were evaluated in terms of their ability to predict self-paced reading times and eye-gaze durations.
The regression analysis revealed a strictly monotonic, positive log-linear relationship between perplexity and fit to reading times for five GPT-Neo variants and eight OPT variants, providing robust empirical support for this trend.
Additionally, the different data used to train each LM family seem to influence the quality of surprisal estimates, although more pre-training data did not necessarily result in surprisal estimates that are more predictive of reading times.

Subsequently, to identify the data points that are responsible for the positive relationship, a post-hoc analysis of the residual errors from each regression model was conducted.
The results showed that the difference in MSEs between regression models containing surprisal predictors from different LM variants was especially large on nouns and adjectives, such as named entity terms and predicative adjectives.
A further inspection of their predictions showed that the trend of MSEs on these words was driven mainly by underpredictions of reading time delays, which were exacerbated as the larger LM variants predicted the words more accurately and assigned lower surprisal values.
This tendency also led to higher regression coefficients for surprisal estimates from larger LM variants, which resulted in a systematic overprediction at function words like conjunctions and modals that had similar surprisal estimates across LM variants.

The `more and more superhuman' predictions of larger LM variants observed in this work are consistent with findings from recent analyses of Transformer-based LMs.
For example, a mathematical analysis of Transformers \citep{elhageetal21} showed that a layer of self-attention essentially functions as a lookup table that keeps track of bigram statistics of the input data.
Given this observation, it may be the case that the larger LM variants with more attention heads at their disposal have the capability to learn stronger local associations between tokens.
This possibility was empirically supported from the perspective of memorization by \citet{carlinietal22}, who found that larger variants of the GPT-Neo model returned more sequences verbatim from the pre-training data during greedy decoding.
This behavior may explain why nouns and adjectives showed the strongest effect in the post-hoc analysis; since adjectives and nouns typically have higher type-frequency than verbs or function words, it may be the case that nouns and adjectives that are rarely seen during training are predicted much more faithfully by the larger LM variants with higher model capacity.
Additionally, this also suggests that as these pre-trained LMs continue to get bigger, they will continue to degrade as models of humanlike language comprehension.

The `trained-from-scratch' LMs studied in earlier psycholinguistic modeling work \citep[e.g.][]{goodkindbicknell18, wilcoxetal20} observe a negative relationship between perplexity and fit to reading times.
However, based on regression results following the same protocols as Section \ref{sec:expl}, surprisal estimates from LMs trained in \citet{wilcoxetal20} generally seem to be less predictive of human reading times than those from pre-trained LMs examined in this work.
Given the especially large discrepancy in model size between newly-trained LMs and pre-trained LMs, it may be the case that they capture two distinct regimes in terms of the relationship between LM performance and predictive power of surprisal estimates.
While the results of the current study clearly show that surprisal estimates from smaller pre-trained LM variants are more predictive of reading times, it remains to be seen how much smaller LMs can become before the predictive power of surprisal estimates starts to decrease.
With recently increasing effort in developing efficient NLP models, future work could explore the extent to which e.g.~knowledge distillation techniques \citep{sanhetal19} can result in LMs that are more predictive of humanlike processing difficulty.

Additionally, the importance of being `adequately surprised' at nouns like named entity terms that was identified in the current study may also explain similar recent counterexamples to the trend observed between model perplexity and fit to reading times \citep{ohetal21acl, kuribayashietal21}.
\citet{ohetal21acl} showed that incorporating a character model to estimate word generation probabilities within an incremental left-corner parser resulted in more predictive surprisal estimates compared to those from a baseline parser that treats words as symbols, although at a cost of higher test perplexity.
The character model may be effectively assigning higher surprisal values to these rare words, thereby achieving better fit to human reading times.
The reading times of Japanese text studied in \citet{kuribayashietal21} were measured in larger units (i.e.~\textit{bunsetsu}; roughly equivalent to phrases) than typical English words.
Therefore, the Japanese LMs analyzed in that study are likely to have been trained on and have made predictions on text that has been tokenized into `sub-bunsetsu' tokens, which may have made a different picture emerge from results based on purely word-based LMs of earlier work.

In general, the tendency of pre-trained LM surprisal to underpredict reading times observed in this work is consistent with recent empirical shortcomings of neural LM surprisal.
For example, \citet{vanschijndellinzen21} and \citet{arehallietal22} found that surprisal from neural LMs severely underpredicts the magnitude of garden-path effects demonstrated by human subjects.~Similarly, \citet{hahnetal22} showed that surprisal from GPT-2 fails to accurately predict the increase in reading times at the main verb of deeply embedded sentences.
\citet{kuribayashietal22} also demonstrated that implementing a recency bias by deterministically truncating the context window of neural LMs leads to surprisal estimates that alleviate the underpredictions of full neural LM surprisal on naturalistic reading times of English and Japanese text.
Taken together, these results suggest that neural LMs do not make abstract, linguistic generalizations like people do.

Moreover, there are also efforts to evaluate other memory- and attention-based predictors calculated from Transformer-based LM representations on their ability to predict human behavior.
For instance, \citet{ryulewis21} drew connections between the self-attention mechanism of Transformers and cue-based retrieval models of sentence comprehension \citep[e.g.][]{lewisetal06}.
Their proposed attention entropy, which quantifies the diffuseness of attention weights over previous tokens, was found to show profiles that are consistent with similarity-based interference observed during the processing of subject-verb agreement.
\citet{ohschuler22} expanded upon this idea and showed that the entropy of attention weights at a given timestep as well as the shift in attention weights across consecutive timesteps are robust predictors of naturalistic reading times over GPT-2 surprisal.
\citet{hollensteinbeinborn21} calculated the norm of the gradient of each input token on two eye-tracking corpora using BERT \citep{devlinetal19} as a metric of saliency, which showed higher correlations to fixation durations compared to raw attention weights.

Finally, it is becoming more common in psycholinguistic modeling to use surprisal from pre-trained LMs as a baseline predictor to study various effects in naturalistic sentence processing \citep[e.g.][]{ryulewis22, clarkschuler22}.
The broader implication of the current study is that researchers should not select the largest pre-trained LM available based on the widely-held `larger is better' assumption of the NLP community.
As a general practice, surprisal estimates from smaller pre-trained LM variants should be incorporated to form a more rigorous baseline, which will guard against drawing unwarranted scientific conclusions.
\section*{Acknowledgments}
We thank TACL action editor Marco Baroni and the reviewers for their helpful comments.
This work was supported by the National Science Foundation grant \#1816891.
All views expressed are those of the authors and do not necessarily reflect the views of the National Science Foundation.
\bibliography{tacl2021}
\bibliographystyle{acl_natbib}








  

\end{document}